\documentclass[10pt,twocolumn,letterpaper]{article}

\usepackage{cvpr}
\usepackage{times}
\usepackage{epsfig}
\usepackage{graphicx}
\usepackage{amsmath}
\usepackage{amssymb}
\usepackage{mathtools}
\usepackage{calligra}
\usepackage{color}
\usepackage{array}
\usepackage{hhline}
\usepackage{subfig}
\usepackage{multirow}
\usepackage{caption}
\usepackage{boxhandler}
\usepackage{tabularx}
\usepackage{adjustbox}
\usepackage{booktabs}
\usepackage{array}
\usepackage{overpic}
\usepackage{bbding}
\newcolumntype{C}{>{\centering\arraybackslash}X}
\usepackage[justification=centering]{caption}

\newcolumntype{P}[1]{>{\centering\arraybackslash}p{#1}}
\newcolumntype{M}[1]{>{\centering\arraybackslash}m{#1}}

\cvprfinalcopy 

\def\cvprPaperID{916} 

\ifcvprfinal\fi
\begin{document}

\title{Toward Convolutional Blind Denoising of Real Photographs}

\author{Shi Guo$^{1,3,4}$, Zifei Yan$^{(}$\Envelope$^{)}$ $^{1}$, Kai Zhang$^{1,3}$, Wangmeng Zuo$^{1,2}$, Lei Zhang$^{3,4}$\\
$^1$Harbin Institute of Technology, Harbin; $^2$Peng Cheng Laboratory, Shenzhen; \\
 $^3$ The Hong Kong Polytechnic University, Hong Kong; $^4$DAMO Academy, Alibaba Group\\
{\tt\small guoshi28@outlook.com, \{wmzuo,yanzifei\}@hit.edu.cn}\\
{\tt\small cskaizhang@gmail.com, cslzhang@comp.polyu.edu.hk}
}


\maketitle
\thispagestyle{empty}

\begin{abstract}
%
%
%
%
%
%
%
While deep convolutional neural networks (CNNs) have achieved impressive success in image denoising with additive white Gaussian noise (AWGN), their performance remains limited on real-world noisy photographs.
The main reason is that their learned models are easy to overfit on the simplified AWGN model which deviates severely from the complicated real-world noise model.
In order to improve the generalization ability of deep CNN denoisers, we suggest training a convolutional blind denoising network (CBDNet) with more realistic noise model and real-world noisy-clean image pairs.
On the one hand, both signal-dependent noise and in-camera signal processing pipeline is considered to synthesize realistic noisy images.
On the other hand, real-world noisy photographs and their nearly noise-free counterparts are also included to train our CBDNet.
To further provide an interactive strategy to rectify denoising result conveniently, a noise estimation subnetwork with asymmetric learning to suppress under-estimation of noise level is embedded into CBDNet.
Extensive experimental results on three datasets of real-world noisy photographs clearly demonstrate the superior performance of CBDNet over state-of-the-arts in terms of quantitative metrics and visual quality.
The code has been made available at \url{https://github.com/GuoShi28/CBDNet}.


\end{abstract}

\section{Introduction}
Image denoising is an essential and fundamental problem in low-level vision and image processing.
With decades of studies, numerous promising approaches~\cite{mention_Aharon2006KSVDAA,compare_Dabov2007ColorID,compare_Gu2014WeightedNN,mention_Schmidt2014ShrinkageFF,mention_Chen2017TrainableNR,mention_Zhang2017BeyondAG} have been developed and near-optimal performance~\cite{mention_Chatterjee2010IsDD,mention_Levin2012PatchCF,mention_Romano2016TheLE} has been achieved for the removal of additive white Gaussian noise (AWGN).
However, in real camera system, image noise comes from multiple sources (\eg, dark current noise, short noise, and thermal noise) and is further affected by in-camera processing (ISP) pipeline (\eg, demosaicing, Gamma correction, and compression).
All these make real noise much more different from AWGN, and blind denoising of real-world noisy photographs remains a challenging issue.

	\begin{figure}[t]
	\setlength{\abovecaptionskip}{0.cm}
		\begin{center}
			\newcommand{\rowArg}{2.605cm}
			\newcommand{\fullSize}{5.626cm}
			\newcommand{\patchSize}{2.55cm}
			\setlength\tabcolsep{0.05cm}
			\begin{tabular}[b]{c c c}
				\multicolumn{2}{c}{\multirow{2}{*}[\rowArg]{
						\subfloat[``0002\_02'' from DND~\cite{dataset_Plotz2017BenchmarkingDA}]
						{\includegraphics[height=\fullSize, trim={0.55cm 0cm 0.55cm 0cm}, clip]
							{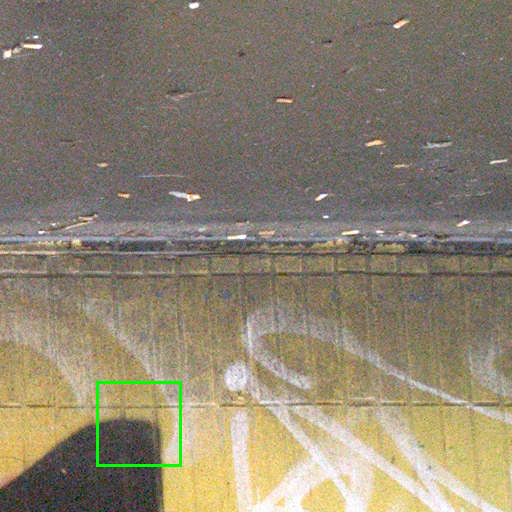}}}} &
				\subfloat[Noisy]{
					\includegraphics[width = \patchSize, height = \patchSize]
					{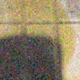}} \\ [-0.3cm]& &
				\subfloat[BM3D~\cite{compare_Dabov2007ColorID}]{
					\includegraphics[width = \patchSize, height = \patchSize]
					{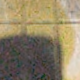}} \\ [-0.3cm]
				
				\subfloat[DnCNN~\cite{mention_Zhang2017BeyondAG}]{
					\includegraphics[width = \patchSize, height = \patchSize]
					{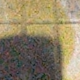}} &
				\subfloat[FFDNet+~\cite{mention_Zhang2017FFDNetTA}]{
					\includegraphics[width = \patchSize, height = \patchSize]
					{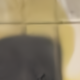}}	&
				\subfloat[\textbf{CBDNet}]{
					\includegraphics[width = \patchSize, height = \patchSize]
					{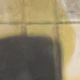}}		
			\end{tabular}		
		\end{center}
		\setlength{\abovecaptionskip}{-0.2cm}
		\captionsetup{justification=raggedright,singlelinecheck=false}
		\caption{Denoising results of different methods on real- world noisy image ``0002\_02'' from DND~\cite{dataset_Plotz2017BenchmarkingDA}.}
		\label{fig1}\vspace{-0.8cm}
	\end{figure}

In the recent past, Gaussian denoising performance has been significantly advanced by the development of deep CNNs~\cite{mention_Zhang2017BeyondAG,mention_Mao2016ImageRU,mention_Zhang2017FFDNetTA}.
However, deep denoisers for blind AWGN removal degrades dramatically when applied to real photographs (see Fig.~\ref{fig1}(d)). On the other hand, deep denoisers for non-blind AWGN removal would smooth out the details while removing the noise (see Fig.~\ref{fig1}(e)).
%
%
Such an phenomenon may be explained from the characteristic of deep CNNs~\cite{mention_Martin2017RethinkingGR}, where their generalization largely depends on the ability of memorizing large scale training data.
In other words, existing CNN denoisers tend to be over-fitted to Gaussian noise and generalize poorly to real-world noisy images with more sophisticated noise.


In this paper, we tackle this issue by developing a convolutional blind denoising network (CBDNet) for real-world photographs.
As indicated by~\cite{mention_Martin2017RethinkingGR}, the success of CNN denoisers are significantly dependent on whether the distributions of synthetic and real noises are well matched.
Therefore, realistic noise model is the foremost issue for blind denoising of real photographs.
According to~\cite{mention_Foi2008PracticalPN,dataset_Plotz2017BenchmarkingDA}, Poisson-Gaussian distribution which can be approximated as heteroscedastic Gaussian of a signal-dependent and a stationary noise components has been considered as a more appropriate alternative than AWGN for real raw noise modeling.
Moreover, in-camera processing would further makes the noise spatially and chromatically correlated which increases the complexity of noise.
As such, we take into account both Poisson-Gaussian model and in-camera processing pipeline (\eg, demosaicing, Gamma correction, and JPEG compression) in our noise model.
Experiments show that in-camera processing pipeline plays a pivot role in realistic noise modeling, and achieves notably performance gain (\ie, $> 5$ dB by PSNR) over AWGN on DND~\cite{dataset_Plotz2017BenchmarkingDA}.

We further incorporate both synthetic and real noisy images to train CBDNet.
On one hand, it is easy to access massive synthetic noisy images.
However, the noise in real photographs cannot be fully characterized by our model, thereby giving some leeway for improving denoising performance.
On the other hand, several approaches~\cite{mention_Nam2016AHA,abdelhamed2018high} have suggested to get noise-free image by averaging hundreds of noisy images at the same scene.
Such solution, however, is expensive in cost, and suffers from the over-smoothing effect of noise-free image.
Benefited from the incorporation of synthetic and real noisy images, $0.3\sim0.5$ dB gain on PSNR can be attained by CBDNet on DND~\cite{dataset_Plotz2017BenchmarkingDA}.

Our CBDNet is comprised of two subnetworks, \ie, noise estimation and non-blind denoising. %
With the introduction of noise estimation subnetwork, we adopt an asymmetric loss by imposing more penalty on under-estimation error of noise level, making our CBDNet perform robustly when the noise model is not well matched with real-world noise.
Besides, it also allows the user to interactively rectify the denoising result by tuning the estimated noise level map.
Extensive experiments are conducted on three real noisy image datasets, \ie, NC12~\cite{dataset_Lebrun2015TheNC}, DND~\cite{dataset_Plotz2017BenchmarkingDA} and Nam~\cite{mention_Nam2016AHA}.
In terms of both quantitative metrics and perceptual quality, our CBDNet performs favorably in comparison to state-of-the-arts.
As shown in Fig.~\ref{fig1}, both non-blind BM3D~\cite{compare_Dabov2007ColorID} and DnCNN for blind AWGN~\cite{mention_Zhang2017BeyondAG} fail to denoise the real-world noisy photograph.
In contrast, our CBDNet achieves very pleasing denoising results by retaining most structure and details while removing the sophisticated real-world noise.

To sum up, the contribution of this work is four-fold:

\begin{itemize}
  \item A realistic noise model is presented by considering both heteroscedastic Gaussian noise and in-camera processing pipeline, greatly benefiting the denoising performance.
  \item Synthetic noisy images and real noisy photographs are incorporated for better characterizing real-world image noise and improving denoising performance.
  \item Benefited from the introduction of noise estimation subnetwork, asymmetric loss is suggested to improve the generalization ability to real noise, and interactive denoising is allowed by adjusting the noise level map.
  \item Experiments on three real-world noisy image datasets show that our CBDNet achieves state-of-the-art results in terms of both quantitative metrics and visual quality.
\end{itemize}


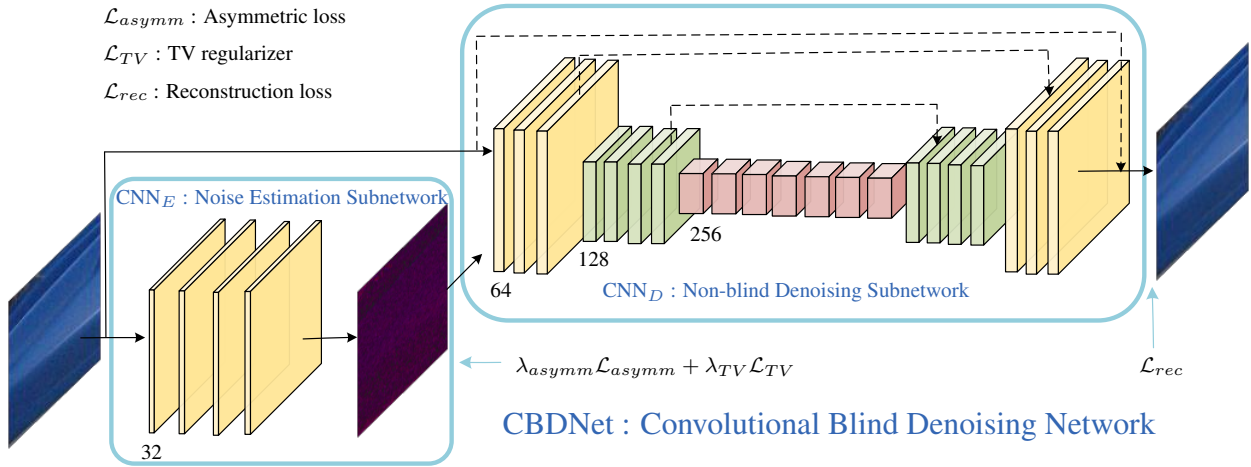
\begin{figure*}[!t]
\centering
	\begin{overpic}[width=0.95\textwidth]{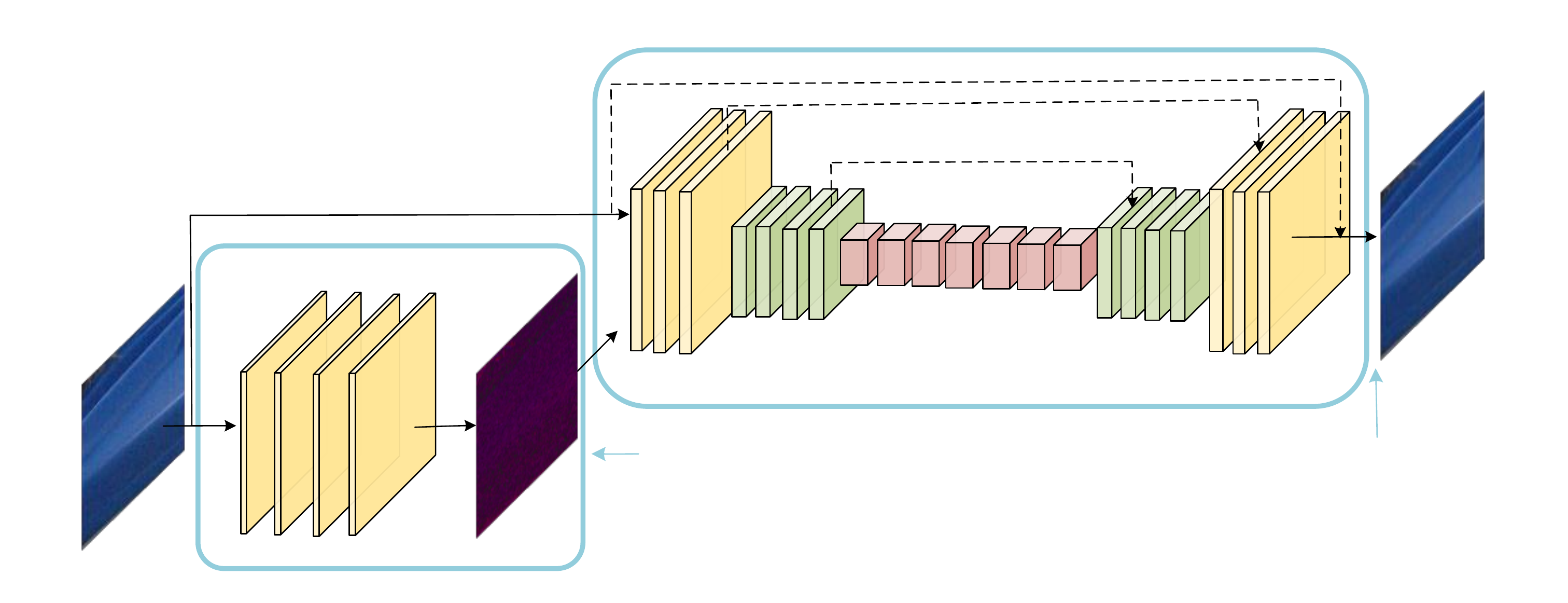}
		\put(8,37){\color{black}{\footnotesize $\mathcal{L}_{asymm}$ : Asymmetric loss}}
		\put(8,34){\color{black}{\footnotesize $\mathcal{L}_{TV}$ : TV regularizer}}
		\put(8,31){\color{black}{\footnotesize $\mathcal{L}_{rec}$ : Reconstruction loss}}
		\put(11,2){\color{black}{\footnotesize 32}}
		\put(41,9){\color{black}{\footnotesize $\lambda_{asymm} \mathcal{L}_{asymm} + \lambda_{TV} \mathcal{L}_{TV}$}}
		\put(91,9){\color{black}{\footnotesize $\mathcal{L}_{rec}$}}
		\put(39,15){\color{black}{\footnotesize 64}}
		\put(46,17.5){\color{black}{\footnotesize 128}}
		\put(55,19.5){\color{black}{\footnotesize 256}}
		\put(9,22.5){\textcolor[rgb]{0.2,0.4,0.7}{\footnotesize $\text{CNN}_E$ : Noise Estimation Subnetwork}}
		\put(48,15){\textcolor[rgb]{0.2,0.4,0.7}{\footnotesize $\text{CNN}_D$ : Non-blind Denoising Subnetwork}}
		\put(40,4){\textcolor[rgb]{0.2,0.4,0.7}{\large CBDNet : Convolutional Blind Denoising Network}}
	\end{overpic}
\caption{Illustration of our CBDNet for blind denoising of real-world noisy photograph.}
\label{figFlow}
\vspace{-0.5cm}
\end{figure*}

\vspace{-0.2cm}
\section{Related Work}
\subsection{Deep CNN Denoisers}
The advent of deep neural networks (DNNs) has led to great improvement on Gaussian denoising.
Until Burger \etal \cite{mention_Burger2012ImageDC}, most early deep models cannot achieve state-of-the-art denoising performance~\cite{mention_Jain2008NaturalID,mention_Rabie2005RobustEA,mention_Xie2012ImageDA}.
Subsequently, CSF~\cite{mention_Schmidt2014ShrinkageFF} and TNRD \cite{mention_Chen2017TrainableNR} unroll the optimization algorithms for solving the fields of experts model to learn stage-wise inference procedure.
By incorporating residual learning \cite{mention_He2016DeepRL} and batch normalization \cite{mention_Ioffe2015BatchNA}, Zhang \etal \cite{mention_Zhang2017BeyondAG} suggest a denoising CNN (DnCNN) which can outperform traditional non-CNN based methods.
Without using clean data, Noise2Noise~\cite{lehtinen2018noise2noise} also achieves state-of-the-art.
Most recently, other CNN methods, such as RED30~\cite{mention_Mao2016ImageRU}, MemNet~\cite{mention_Tai2017MemNetAP}, BM3D-Net~\cite{Yang2018BM3DNetAC}, MWCNN~\cite{Liu2018MultilevelWF} and FFDNet~\cite{mention_Zhang2017FFDNetTA}, are also developed with promising denoising performance.

Benefited from the modeling capability of CNNs, the studies~\cite{mention_Zhang2017BeyondAG,mention_Mao2016ImageRU,mention_Tai2017MemNetAP} show that it is feasible to learn a single model for blind Gaussian denoising.
However, these blind models may be over-fitted to AWGN and fail to handle real noise.
In contrast, non-blind CNN denoisiers, \eg, FFDNet~\cite{mention_Zhang2017FFDNetTA}, can achieve satisfying results on most real noisy images by manually setting proper or relatively higher noise level.
To exploit this characteristic, our CBDNet includes a noise estimation subnetwork as well as an asymmetric loss to suppress under-estimation error of noise level.

\subsection{Image Noise Modeling}
Most denoising methods are developed for non-blind Gaussian denoising.
However, the noise in real images comes from various sources (dark current noise, short noise, thermal noise, etc.), and is much more sophisticated~\cite{mention_Ortiz2004RadiometricCO}.
By modeling photon sensing with Poisson and remaining stationary disturbances with Gaussian, Poisson-Gaussian noise model \cite{mention_Foi2008PracticalPN} has been adopted for the raw data of imaging sensors.
In \cite{mention_Foi2008PracticalPN,Liu2008AutomaticEA}, camera response function (CRF) and quantization noise are also considered for more practical noise modeling.
Instead of Poisson-Gaussian, Hwang \etal~\cite{mention_Hwang2012DifferenceBasedIN} present a Skellam distribution for Poisson photon noise modeling.
Moreover, when taking in-camera image processing pipeline into account, the channel-independent noise assumption may not hold true, and several approaches \cite{mention_Kim2012ANI,mention_Nam2016AHA} are proposed for cross-channel noise modeling.
In this work, we show that realistic noise model plays a pivot role in CNN-based denoising of real photographs, and both Poisson-Gaussian noise and in-camera image processing pipeline benefit denoising performance.

\subsection{Blind Denoising of Real Images}
Blind denoising of real noisy images generally is more challenging and can involve two stages, \ie, noise estimation and non-blind denoising.
For AWGN, several PCA-based \cite{mention_Pyatykh2013ImageNL,mention_Liu2013SingleImageNL,Chen2015AnES} methods have been developed for estimating noise standard deviation ($SD.$).
Rabie \cite{mention_Rabie2005RobustEA} models the noisy pixels as outliers and exploits Lorentzian robust estimator for AWGN estimation.
For Poisson-Gaussian model, Foi \etal \cite{mention_Foi2008PracticalPN} suggest a two-stage scheme, \ie, local estimation of multiple expectation/standard-deviation pairs, and global parametric model fitting.

In most blind denoising methods, noise estimation is closely coupled with non-blind denoising.
Portilla \cite{mention_Portilla2004BlindNN,mention_Portilla2004FullBD} adopts a Gaussian scale mixture for modeling wavelet patches of each scale, and utilizes Bayesian least square to estimate clean wavelet patches.
Based on the piecewise smooth image model, Liu \etal \cite{Liu2008AutomaticEA} propose a unified framework for the estimation and removal of color noise.
%
Gong \etal \cite{mention_Gong2014ImageRW} model the data fitting term as the weighted sum of the $L_1$ and $L_2$ norms, and utilize a sparsity regularizer in wavelet domain for handling mixed or unknown noises.
Lebrun \etal \cite{compare_Lebrun2015MultiscaleIB,dataset_Lebrun2015TheNC} propose an extension of non-local Bayes approach \cite{mention_lebrun2013nonlocal} by modeling the noise of each patch group to be zero-mean correlated Gaussian distributed.
Zhu \etal \cite{mention_Zhu2016FromNM} suggest a Bayesian nonparametric technique to remove the noise via the low-rank mixture of Gaussians (LR-MoG) model.
Nam \etal \cite{mention_Nam2016AHA} model the cross-channel noise as a multivariate Gaussian and perform denoising by the Bayesian nonlocal means filter \cite{mention_Kervrann2007BayesianNM}.
Xu \etal~\cite{Xu2017MultichannelWN} suggest a multi-channel weighted nuclear norm minimization (MCWNNM) model to exploit channel redundancy.
They further present a trilateral weighted sparse coding (TWSC) method for better modeling noise and image priors~\cite{Xu2018ATW}.
%
%
Except noise clinic (NC) \cite{compare_Lebrun2015MultiscaleIB,dataset_Lebrun2015TheNC}, MCWNNM~\cite{Xu2017MultichannelWN}, and TWSC~\cite{Xu2018ATW}, the codes of most blind denoisers are not available.
Our experiments show that they are still limited for removing noise from real images.

\section{Proposed Method}
This section presents our CBDNet consisting of a noise estimation subnetwork and a non-blind denoising subnetwork. 
To begin with, we introduce the noise model to generate synthetic noisy images.
Then, the network architecture and asymmetric loss.
Finally, we explain the incorporation of synthetic and real noisy images for training CBDNet.

\subsection{Realistic Noise Model}\label{secnoisemodel}
As noted in~\cite{mention_Martin2017RethinkingGR}, the generalization of CNN largely depends on the ability in memorizing training data.
Existing CNN denoisers, \eg, DnCNN~\cite{mention_Zhang2017BeyondAG}, generally does not work well on real noisy images, mainly due to that they may be over-fitted to AWGN while the real noise distribution is much different from Gaussian.
On the other hand, when trained with a realistic noise model, the memorization ability of CNN will be helpful to make the learned model generalize well to real photographs.
Thus, noise model plays a critical role in guaranteeing performance of CNN denoiser.

Different from AWGN, real image noise generally is more sophisticated and signal-dependent~\cite{Liu2014PracticalSN,mention_Foi2008PracticalPN}.
Practically, the noise produced by photon sensing can be modeled as Poisson, while the remaining stationary disturbances can be modeled as Gaussian.
Poisson-Gaussian thus provides a reasonable noise model for the raw data of imaging sensors~\cite{mention_Foi2008PracticalPN}, and can be further approximated with a heteroscedastic Gaussian $\mathbf{n}(\mathbf{L}) \sim \mathcal{N}(0, \sigma^2(\mathbf{L}))$ defined as,

\small
\begin{equation}
\sigma^2(\mathbf{L}) = \mathbf{L} \cdot \sigma_s^2 + \sigma_c^2.
\label{noiselevelmap}
\end{equation}
\normalsize
where $\mathbf{L}$ is the irradiance image of raw pixels.
$\mathbf{n}(\mathbf{L}) = \mathbf{n}_s(\mathbf{L}) + \mathbf{n}_c$ involves two components, \ie, a stationary noise component $\mathbf{n}_c$ with noise variance ${\sigma}_c^2$ and a signal-dependent noise component $\mathbf{n}_s$ with spatially variant noise variance $\mathbf{L} \cdot {\sigma}_s^2$.

Real photographs, however, are usually obtained after in-camera processing (ISP), which further increases the complexity of noise and makes it spatially and chromatically correlated.
Thus, we take two main steps of ISP pipeline, \ie, demosaicing and Gamma correction, into consideration, resulting in the realistic noise model as,
\small
\begin{equation}
\mathbf{y} = f(\mathbf{DM}(\mathbf{L} + \mathbf{n}(\mathbf{L}) )),
\label{noisesynthsismodel}
\end{equation}
\normalsize
where $\mathbf{y}$ denotes the synthetic noisy image, $f(\cdot)$ stands for the camera response function (CRF) uniformly sampled from the 201 CRFs provided in \cite{dataset_Grossberg2004ModelingTS}.
And $\mathbf{L} = \mathbf{M}f^{-1}(\mathbf{x})$ is adopted to generate irradiance image from a clean image $\mathbf{x}$.
$\mathbf{M}(\cdot)$ represents the function that converts sRGB image to Bayer image and $\mathbf{DM}(\cdot)$ represents the demosaicing function \cite{mention_malvar2004high}.
Note that the interpolation in $\mathbf{DM}(\cdot)$ involves pixels of different channels and spatial locations.
The synthetic noise in Eqn.~(\ref{noisesynthsismodel}) is thus channel and space dependent.

Furthermore, to extend CBDNet for handling compressed image, we can include JPEG compression in generating synthetic noisy image,
\small
\begin{equation}
\mathbf{y} = JPEG(f(\mathbf{DM}(\mathbf{L} + \mathbf{n}(\mathbf{L})))).
\label{noisesynthsismodeljpeg}
\end{equation}
\normalsize
%
%
For noisy uncompressed image, we adopt the model in Eqn.~(\ref{noisesynthsismodel}) to generate synthetic noisy images.
For noisy compressed image, we exploit the model in Eqn.~(\ref{noisesynthsismodeljpeg}).
Specifically, $\sigma_s$ and $\sigma_c$ are uniformly sampled from the ranges of $[0, 0.16]$ and $[0, 0.06]$, respectively.
In JPEG compression, the quality factor is sampled from the range $[60, 100]$.
We note that the quantization noise is not considered because it is minimal and can be ignored without any obvious effect on denoising result~\cite{mention_Zhang2017FFDNetTA}.

\subsection{Network Architecture}
\label{network structure}
As illustrated in Fig.~\ref{figFlow}, the proposed CBDNet includes a noise estimation subnetwork $\text{CNN}_{E}$ and a non-blind denosing subnetwork $\text{CNN}_{D}$.
First, $\text{CNN}_{E}$ takes a noisy observation $\mathbf{y}$ to produce the estimated noise level map $\hat{\sigma}(\mathbf{y}) = \mathcal{F}_E(\mathbf{y}; \mathbf{W}_E)$, where $\mathbf{W}_E$ denotes the network parameters of $\text{CNN}_{E}$.
We let the output of $\text{CNN}_{E}$ be the noise level map due to that it is of the same size with the input $\mathbf{y}$ and can be estimated with a fully convolutional network.
Then, $\text{CNN}_{D}$ takes both $\mathbf{y}$ and $\hat{\sigma}(\mathbf{y})$ as input to obtain the final denoising result $\hat{\mathbf{x}} = \mathcal{F}_D(\mathbf{y}, \hat{\sigma}(\mathbf{y}); \mathbf{W}_D)$, where $\mathbf{W}_D$ denotes the network parameters of $\text{CNN}_{D}$.
Moreover, the introduction of $\text{CNN}_{E}$ also allows us to adjust the estimated noise level map $\hat{\sigma}(\mathbf{y})$ before putting it to the the non-blind denosing subnetwork $\text{CNN}_{D}$.
In this work, we present a simple strategy by letting $\hat{\varrho}(\mathbf{y}) = \gamma \cdot \hat{\sigma}(\mathbf{y})$ for interactive denoising.

We further explain the network structures of $\text{CNN}_{E}$ and $\text{CNN}_{D}$.
%
$\text{CNN}_{E}$ adopts a plain five-layer fully convolutional network without pooling and batch normalization operations.
In each convolution (Conv) layer, the number of feature channels is set as $32$, and the filter size is $3 \times 3$.
The ReLU nonlinearity~\cite{mention_Nair2010RectifiedLU} is deployed after each Conv layer.
As for $\text{CNN}_{D}$, we adopt an U-Net \cite{ronneberger2015u} architecture which takes both $\mathbf{y}$ and $\hat{\sigma}(\mathbf{y})$ as input to give a prediction $\hat{\mathbf{x}}$ of the noise-free clean image.
Following~\cite{mention_Zhang2017BeyondAG}, the residual learning is adopted by first learning the residual mapping $\mathcal{R}(\mathbf{y}, \hat{\sigma}(\mathbf{y}); \mathbf{W}_D)$ and then predicting $\hat{\mathbf{x}} = \mathbf{y} + \mathcal{R}(\mathbf{y}, \hat{\sigma}(\mathbf{y}); \mathbf{W}_D)$.
%
The 16-layer U-Net architecture of $\text{CNN}_{E}$ is also given in Fig.~\ref{figFlow}, where symmetric skip connections, strided convolutions and transpose convolutions are introduced for exploiting multi-scale information as well as enlarging receptive field.
All the filter size is $3 \times 3$, and the ReLU nonlinearity~\cite{mention_Nair2010RectifiedLU} is applied after every Conv layer except the last one.
Moreover, we empirically find that batch normalization helps little for the noise removal of real photographs, partially due to that the real noise distribution is fundamentally different from Gaussian.

Finally, we note that it is also possible to train a single blind CNN denoiser by learning a direct mapping from noisy observation to clean image.
However, as noted in \cite{mention_Zhang2017FFDNetTA,mildenhall2018burst}, taking both noisy image and noise level map as input is helpful in generalizing the learned model to images beyond the noise model and thus benefits blind denoising.
We empirically find that single blind CNN denoiser performs on par with CBDNet for images with lower noise level, and is inferior to CBDNet for images with heavy noise.
Furthermore, the introduction of noise estimation subnetwork also makes interactive denoising and asymmetric learning allowable.
Therefore, we suggest to include the noise estimation subnetwork in our CBDNet.

%
%
%
%
\subsection{Asymmetric Loss and Model Objective}
\label{objective}
Both CNN and traditional non-blind denoisers perform robustly when the input noise $SD.$ is higher than the ground-truth one (\ie, over-estimation error), which encourages us to adopt asymmetric loss for improving generalization ability of CBDNet.
%
%
As illustrates in FFDNet~\cite{mention_Zhang2017FFDNetTA}, BM3D/FFDNet achieve the best result when the input noise $SD.$ and ground-truth noise $SD.$ are matched.
When the input noise $SD.$ is lower than the ground-truth one, the results of BM3D/FFDNet contain perceptible noises.
When the input noise $SD.$ is higher than the ground-truth one, BM3D/FFDNet can still achieve satisfying results by gradually wiping out some low contrast structure along with the increase of input noise $SD$.
Thus, non-blind denoisers are sensitive to under-estimation error of noise $SD.$, but are robust to over-estimation error.
With such property, BM3D/FFDNnet can be used to denoise real photographs by setting relatively higher input noise $SD.$, and this might explain the reasonable performance of BM3D on the DND benchmark~\cite{dataset_Plotz2017BenchmarkingDA} in the non-blind setting.

To exploit the asymmetric sensitivity in blind denoising, we present an asymmetric loss on noise estimation to avoid the occurrence of under-estimation error on the noise level map.
Given the estimated noise level $\hat{\sigma}({y}_i)$ at pixel $i$ and the ground-truth $\sigma({y}_i)$, more penalty should be imposed to their MSE when $\hat{\sigma}({y}_i) < \sigma({y}_i)$.
Thus, we define the asymmetric loss on the noise estimation subnetwork as,
\small
\begin {equation}
\mathcal{L}_{asymm} = \sum_{i} \vert \alpha - \mathbb{I}_{(\hat{\sigma}({y}_i) - \sigma({y}_i)) < 0} \vert \cdot \left( \hat{\sigma}({y}_i) - \sigma({y}_i) \right)^2 \, ,
\label {eqamse}
\end {equation}
\normalsize
where $\mathbb{I}_e = 1$ for $e < 0$ and 0 otherwise.
By setting $0 < \alpha < 0.5$, we can impose more penalty to under-estimation error to make the model generalize well to real noise.

Furthermore, we introduce a total variation (TV) regularizer to constrain the smoothness of $\hat{\sigma}(\mathbf{y})$,
\small
\begin {equation}
\mathcal{L}_{TV} = \Vert \nabla_h \hat{\sigma}(\mathbf{y}) \Vert_2^2 + \Vert \nabla_v \hat{\sigma}(\mathbf{y}) \Vert_2^2 \, ,
\end {equation}
\normalsize
where $\nabla_h$ ($\nabla_v$) denotes the gradient operator along the horizontal (vertical) direction.
For the output $\hat{\mathbf{x}}$ of non-blind denoising, we define the reconstruction loss as,
\small
\begin {equation}
\mathcal{L}_{rec} = \Vert \hat{\mathbf{x}} - {\mathbf{x}} \Vert_2^2 \, .
\end {equation}
\normalsize
To sum up, the overall objective of our CBDNet is,
\small
\begin {equation}
\mathcal{L} = \mathcal{L}_{rec} + \lambda_{asymm} \mathcal{L}_{asymm} + \lambda_{TV} \mathcal{L}_{TV},
\label{eqobjective}
\end {equation}
\normalsize
where $\lambda_{asymm}$ and $\lambda_{TV}$ denote the tradeoff parameters for the asymmetric loss and TV regularizer, respectively.
In our experiments, the PSNR/SSIM results of CBDNet are reported by minimizing the above objective.
As for qualitative evaluation of visual quality, we train CBDNet by further adding perceptual loss~\cite{johnson2016perceptual} on \texttt{relu3\_3} of VGG-16~\cite{simonyan2014very} to the objective in Eqn.~(\ref{eqobjective}).

\subsection{Training with Synthetic and Real Noisy Images}\label{sec:incorporating_train}
The noise model in Sec.~\ref{secnoisemodel} can be used to synthesize any amount of noisy images.
And we can also guarantee the high quality of the clean images.
Even though, the noise in real photographs cannot be fully characterized by the noise model.
Fortunately, according to~\cite{mention_Nam2016AHA,dataset_Plotz2017BenchmarkingDA,abdelhamed2018high}, nearly noise-free image can be obtained by averaging hundreds of noisy images from the same scene, and several datasets have been built in literatures.
In this case, the scenes are constrained to be static, and it is generally expensive to acquire hundreds of noisy images.
Moreover, the nearly noise-free image tends to be over-smoothing due to the averaging effect.
Therefore, synthetic and real noisy images can be combined to improve the generalization ability to real photographs.

In this work, we use the noise model in Sec.~\ref{secnoisemodel} to generate the synthetic noisy images, and use 400 images from BSD500 \cite{dataset_MartinFTM01}, 1600 images from Waterloo \cite{dataset_Ma2016WaterlooED}, and 1600 images from MIT-Adobe FiveK dataset~\cite{dataset_fivek} as the training data.
Specifically, we use the RGB image $\textbf{x}$ to synthesize clean raw image $\textbf{L} = \textbf{M}f^{-1}(\textbf{x})$ as a reverse ISP process and use the same $f$ to generate noisy image as Eqns. (\ref{noisesynthsismodel}) or (\ref{noisesynthsismodeljpeg}), where $f$ is a CRF randomly sampled from those in \cite{dataset_Grossberg2004ModelingTS}.
As for real noisy images, we utilize the 120 images from the RENOIR dataset~\cite{dataset_Anaya2014RENOIRA}.
In particular, we alternatingly use the batches of synthetic and real noisy images during training.
For a batch of synthetic images, all the losses in Eqn.~(\ref{eqobjective}) are minimized to update CBDNet.
For a batch of real images, due to the unavailability of ground-truth noise level map, only $\mathcal{L}_{rec}$ and $\mathcal{L}_{TV}$ are considered in training.
We empirically find that such training scheme is effective in improving the visual quality for denoising real photographs.

\section{Experimental Results}

\subsection{Test Datasets}
Three datasets of real-world noisy images, \ie, NC12~\cite{dataset_Lebrun2015TheNC}, DND~\cite{dataset_Plotz2017BenchmarkingDA} and Nam~\cite{mention_Nam2016AHA}, are adopted:


\vspace{-0.4cm}
  \paragraph{NC12} includes 12 noisy images.
        The ground-truth clean images are unavailable, and we only report the denoising results for qualitative evaluation.

\vspace{-0.4cm}
  \paragraph{DND} contains 50 pairs of real noisy images and the corresponding nearly noise-free images.
        %
        %
        Analogous to \cite{dataset_Anaya2014RENOIRA}, the nearly noise-free images are obtained by carefully post-processing of the low-ISO images.
        %
        PSNR/SSIM results are obtained through the online submission system.

\vspace{-0.4cm}
  \paragraph{Nam} contains 11 static scenes and for each scene the nearly noise-free image is the mean image of 500 JPEG noisy images.
        We crop these images into $512 \times 512$ patches and randomly select 25 patches for evaluation.

\subsection{Implementation Details}
%
%
%
The model parameters in Eqn.~(\ref{eqobjective}) are given by $\alpha = 0.3$, $\lambda_1 = 0.5$, and $\lambda_2 = 0.05$.
Note that the noisy images from Nam~\cite{mention_Nam2016AHA} are JPEG compressed, while the noisy images from DND~\cite{dataset_Plotz2017BenchmarkingDA} are uncompressed.
Thus we adopt the noise model in Eqn.~(\ref{noisesynthsismodel}) to train CBDNet for DND and NC12, and the model in Eqn.~(\ref{noisesynthsismodeljpeg}) to train CBDNet(JPEG) for Nam.

To train our CBDNet, we adopt the ADAM \cite{Kingma2014AdamAM} algorithm with $\beta_1$ = 0.9.
The method in \cite{He2015DelvingDI} is adopted for model initialization.
The size of mini-batch is 32 and the size of each patch is $128 \times 128$.
All the models are trained with 40 epochs, where the learning rate for the first 20 epochs is $10^{-3}$, and then the learning rate $5 \times 10^{-4}$ is used to further fine-tune the model.
It takes about three days to train our CBDNet with the MatConvNet package \cite{Vedaldi2015MatConvNetC} on a Nvidia GeForce GTX 1080 Ti GPU.
%
%

\subsection{Comparison with State-of-the-arts}
We consider four blind denoising approaches, \ie, NC~\cite{dataset_Lebrun2015TheNC,compare_Lebrun2015MultiscaleIB}, NI~\cite{compare_NI}, MCWNNM~\cite{Xu2017MultichannelWN} and TWSC~\cite{Xu2018ATW} in our comparison.
NI~\cite{compare_NI} is a commercial software and has been included into Photoshop and Corel PaintShop.
Besides, we also include a blind Gaussian denoising method (\ie, CDnCNN-B \cite{mention_Zhang2017BeyondAG}), and three non-blind denoising methods (\ie, CBM3D \cite{compare_Dabov2007ColorID}, WNNM  \cite{compare_Gu2014WeightedNN}, FFDNet \cite{mention_Zhang2017FFDNetTA}).
When apply non-blind denoiser to real photographs, we exploit~\cite{Chen2015AnES} to estimate the noise $SD.$.

\vspace{-0.5cm}
\paragraph{NC12.}
Fig.~\ref{figin12result2} shows the results of an NC12 images.
All the competing methods are limited in removing noise in the dark region.
In comparison, CBDNet performs favorably in removing noise while preserving salient image structures.

\begin{table}[!tbp]
\setlength{\abovecaptionskip}{0.2cm}
\scriptsize
\centering
\caption{The quantitative results on the DND benchmark.}
\label{tableDNDresults}
\begin{tabularx} {\linewidth} { @{}  p{1.7cm}  p{1.7cm}  p{1.5cm}  X X @{}}
\toprule
Method & Blind/Non-blind  & Denoising on & PSNR & SSIM\\

\midrule
CDnCNN-B~\cite{mention_Zhang2017BeyondAG}  & Blind	&sRGB & 32.43	&0.7900 \\
EPLL~\cite{mention_Zoran2011FromLM}        & Non-blind   & sRGB & 33.51 & 0.8244 \\
TNRD~\cite{mention_Chen2017TrainableNR}       & Non-blind    & sRGB & 33.65 & 0.8306 \\
NCSR~\cite{mention_Dong2013NonlocallyCS}       & Non-blind   & sRGB & 34.05 & 0.8351 \\
MLP~\cite{mention_Burger2012ImageDC}         & Non-blind   & sRGB & 34.23 & 0.8331 \\
FFDNet~\cite{mention_Zhang2017FFDNetTA}	& Non-blind   &sRGB	&34.40 &0.8474 \\
BM3D~\cite{compare_Dabov2007ColorID}      & Non-blind     & sRGB & 34.51 & 0.8507 \\
FoE~\cite{mention_Roth2005FieldsOE}        & Non-blind    & sRGB & 34.62 & 0.8845 \\
WNNM~\cite{compare_Gu2014WeightedNN}    & Non-blind    & sRGB & 34.67 & 0.8646 \\
GCBD~\cite{chen2018image} & Blind & sRGB & 35.58 & 0.9217 \\
CIMM~\cite{compare_Anwar2017ChainingIM}	& Non-blind 	& sRGB  &36.04 &0.9136 \\
KSVD~\cite{mention_Aharon2006KSVDAA}      & Non-blind & sRGB & 36.49 & 0.8978 \\
MCWNNM~\cite{Xu2017MultichannelWN}.       & Blind & sRGB & 37.38 & 0.9294 \\
TWSC~\cite{Xu2018ATW} & Blind & sRGB & 37.94 & 0.9403 \\
\midrule
CBDNet(Syn)      &Blind & sRGB & 37.57 & 0.9360 \\
CBDNet(Real)      &Blind & sRGB & 37.72 & 0.9408 \\
\textbf{CBDNet(All)}        & Blind      & sRGB & \textbf{38.06} & \textbf{0.9421}  \\
\toprule
\end{tabularx}\vspace{-0.45cm}
\end{table}

%

\begin{figure*}[!t]
\setlength{\abovecaptionskip}{0.2cm}
\centering
\subfloat{
\begin{minipage}[t]{0.24\textwidth}
\centering
\includegraphics[width=1\textwidth]{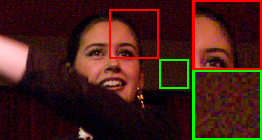}
{\footnotesize  (a) Noisy image}
\end{minipage}\hspace{0.1cm}

\begin{minipage}[t]{0.24\textwidth}
\centering
\includegraphics[width=1\textwidth]{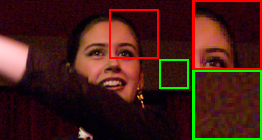}
{\footnotesize  (b) WNNM~\cite{compare_Gu2014WeightedNN}}
\end{minipage}\hspace{0.1cm}

\begin{minipage}[t]{0.24\textwidth}
\centering
\includegraphics[width=1\textwidth]{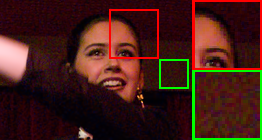}
{\footnotesize  (c) FFDNet~\cite{mention_Zhang2017FFDNetTA}}
\end{minipage}\hspace{0.1cm}

\begin{minipage}[t]{0.24\textwidth}
\centering
\includegraphics[width=1\textwidth]{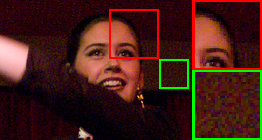}
{\footnotesize  (d) NC~\cite{dataset_Lebrun2015TheNC}}
\end{minipage}
}

\vspace{-3mm}

\subfloat{
\begin{minipage}[t]{0.24\textwidth}
\centering
\includegraphics[width=1\textwidth]{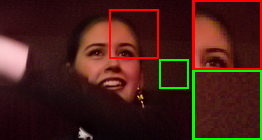}
{\footnotesize  (e) NI~\cite{compare_NI}}
\end{minipage}\hspace{0.1cm}

\begin{minipage}[t]{0.24\textwidth}
\centering
\includegraphics[width=1\textwidth]{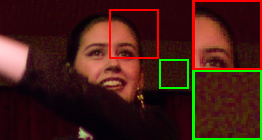}
{\footnotesize  (f) MCWNNM~\cite{Xu2017MultichannelWN}}
\end{minipage}\hspace{0.1cm}

\begin{minipage}[t]{0.24\textwidth}
\centering
\includegraphics[width=1\textwidth]{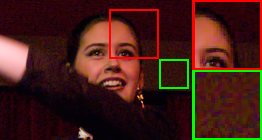}
{\footnotesize  (g) TWSC~\cite{Xu2018ATW}}
\end{minipage}\hspace{0.1cm}

\begin{minipage}[t]{0.24\textwidth}
\centering
\includegraphics[width=1\textwidth]{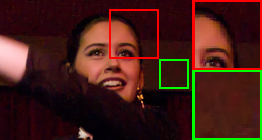}
{\footnotesize  (h) CBDNet}
\end{minipage}
}

\caption{Denoising results of another NC12 image by different methods.}
\label{figin12result2}
\end{figure*}


\begin{figure*}[!t]
\setlength{\abovecaptionskip}{0.2cm}
\centering
\subfloat{
\begin{minipage}[t]{0.24\textwidth}
\centering
\includegraphics[width=1\textwidth]{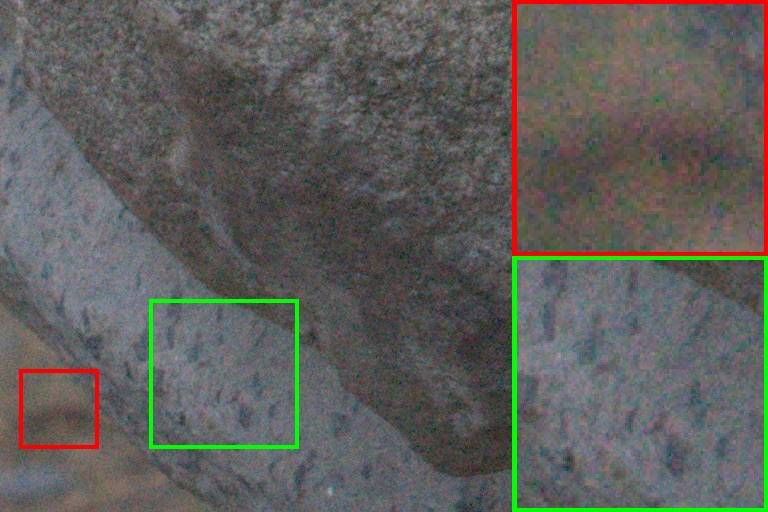}
{\footnotesize  (a) Noisy image}
\end{minipage}\hspace{0.1cm}

\begin{minipage}[t]{0.24\textwidth}
\centering
\includegraphics[width=1\textwidth]{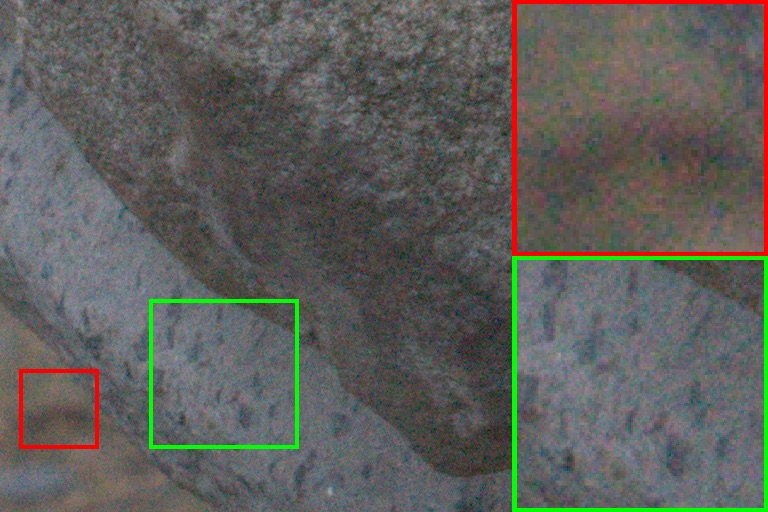}
{\footnotesize  (b) BM3D~\cite{compare_Dabov2007ColorID}}
\end{minipage}\hspace{0.1cm}

\begin{minipage}[t]{0.24\textwidth}
\centering
\includegraphics[width=1\textwidth]{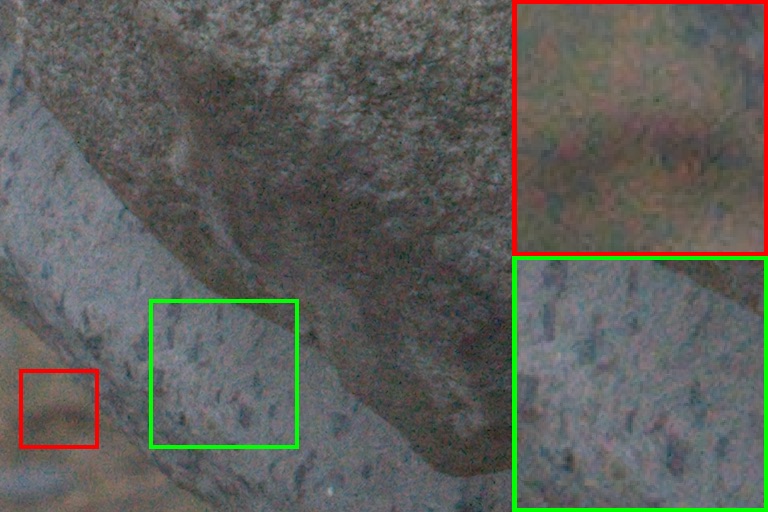}
{\footnotesize  (c) CDnCNN-B~\cite{mention_Zhang2017BeyondAG}}
\end{minipage}\hspace{0.1cm}

\begin{minipage}[t]{0.24\textwidth}
\centering
\includegraphics[width=1\textwidth]{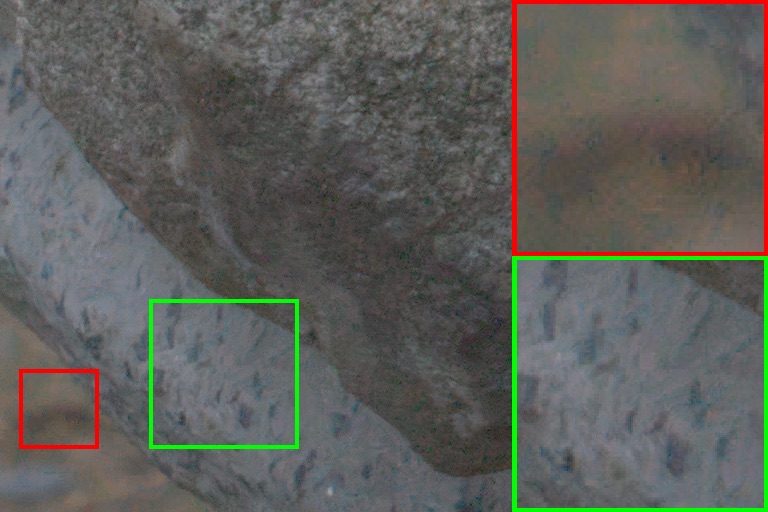}
{\footnotesize  (d) NC~\cite{dataset_Lebrun2015TheNC}}
\end{minipage}

}

\vspace{-3mm}

\subfloat{
\begin{minipage}[t]{0.24\textwidth}
\centering
\includegraphics[width=1\textwidth]{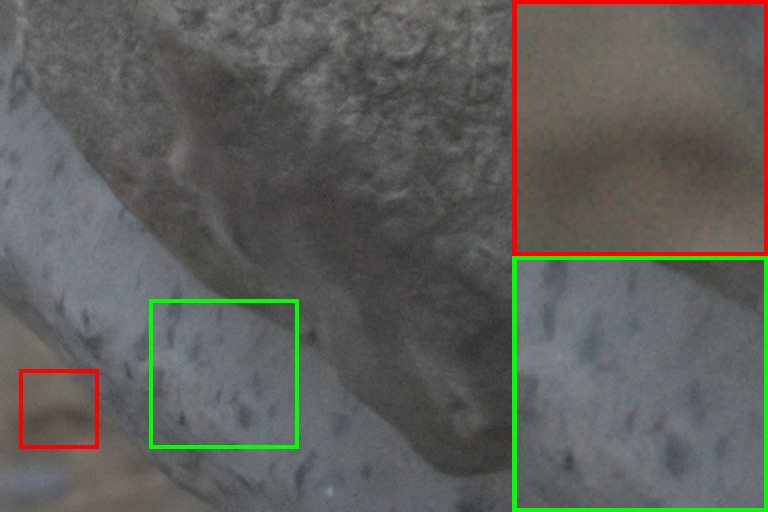}
{\footnotesize  (e) NI~\cite{compare_NI}}
\end{minipage}\hspace{0.1cm}

\begin{minipage}[t]{0.24\textwidth}
\centering
\includegraphics[width=1\textwidth]{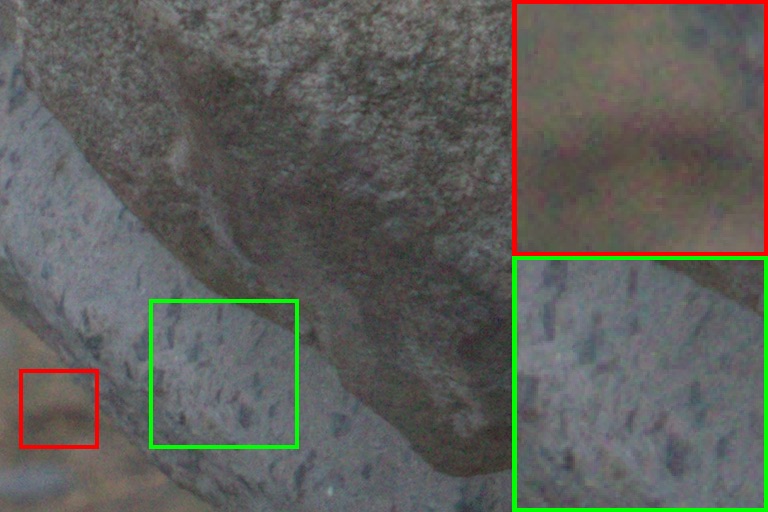}
{\footnotesize  (f) MCWNNM~\cite{Xu2017MultichannelWN}}
\end{minipage}\hspace{0.1cm}

\begin{minipage}[t]{0.24\textwidth}
\centering
\includegraphics[width=1\textwidth]{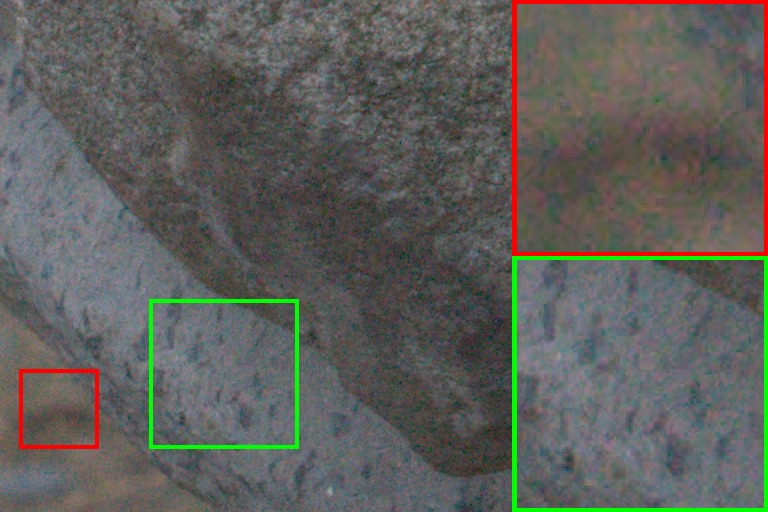}
{\footnotesize  (g) TWSC~\cite{Xu2018ATW}}
\end{minipage}\hspace{0.1cm}

\begin{minipage}[t]{0.24\textwidth}
\centering
\includegraphics[width=1\textwidth]{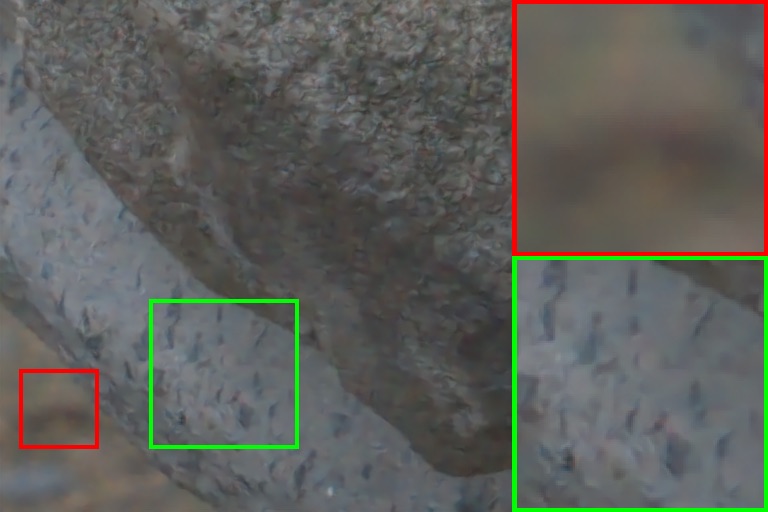}
{\footnotesize  (h) CBDNet}
\end{minipage}
}

\caption{Denoising results of a DND image by different methods.}
\label{figdndresult1}
\end{figure*}

\vspace{-0.5cm}
\paragraph{DND.}
Table~\ref{tableDNDresults} lists the PSNR/SSIM results released on the DND benchmark website.
Undoubtedly, CDnCNN-B~\cite{mention_Zhang2017BeyondAG} cannot be generalized to real noisy photographs and performs very poorly.
Although the noise $SD.$ is provided, non-blind Gaussian denoisers, \eg, WNNM~\cite{compare_Gu2014WeightedNN}, BM3D~\cite{compare_Dabov2007ColorID} and FoE~\cite{mention_Roth2005FieldsOE}, only achieve limited performance, mainly due to that the real noise is much different from AWGN.
MCWNNM~\cite{Xu2017MultichannelWN} and TWSC~\cite{Xu2018ATW} are specially designed for blind denoising of real photographs, and also achieve promising results.
Benefited from the realistic noise model and incorporation with real noisy images, our CBDNet achieves the highest PSNR/SSIM results, and slightly better than MCWNNM~\cite{Xu2017MultichannelWN} and TWSC~\cite{Xu2018ATW}.
CBDNet also significantly outperforms another CNN-based denoiser, \ie, CIMM~\cite{compare_Anwar2017ChainingIM}.
As for running time, CBDNet takes about 0.4s to process an $512 \times 512$ image.
Fig.~\ref{figdndresult1} provides the denoising results of an DND image.
BM3D and CDnCNN-B fail to remove most noise from real photograph,
NC, NI, MCWNNM and TWSC still cannot remove all noise, and NI also suffers from the over-smoothing effect.
In comparison, our CBDNet performs favorably in balancing noise removal and structure preservation.

\begin{table}[!tbp]
\setlength{\abovecaptionskip}{0.2cm}
\scriptsize
\centering

\caption{The quantitative results on the Nam dataset~\cite{mention_Nam2016AHA}.}
\label{tableNamresults}

\begin{tabularx}  {\linewidth} { @{} p{2.0cm}  p{2.0cm} X X @{} }
\toprule
Method       & Blind/Non-blind     &  PSNR      & SSIM   \\
\midrule
NI~\cite{compare_NI}                                  & Blind & 31.52   & 0.9466 \\
CDnCNN-B~\cite{mention_Zhang2017BeyondAG} & Blind & 37.49 & 0.9272 \\
TWSC~\cite{Xu2018ATW}                           & Blind & 37.52  & 0.9292 \\
MCWNNM~\cite{Xu2017MultichannelWN}  & Blind & 37.91  & 0.9322 \\
BM3D~\cite{compare_Dabov2007ColorID} & Non-blind  & 39.84 & 0.9657 \\
NC~\cite{dataset_Lebrun2015TheNC}        & Blind & 40.41  & 0.9731 \\
WNNM~\cite{compare_Gu2014WeightedNN}  & Non-blind & 41.04 & 0.9768 \\

\midrule
CBDNet                                                        & Blind & 40.02  & 0.9687 \\
 \textbf{CBDNet(JPEG)}                               & Blind & \textbf{41.31} & \textbf{0.9784} \\
\toprule
\end{tabularx}
\centering
\vspace{-0.4cm}
\end{table}

\begin{table}[!htbp]
\setlength{\abovecaptionskip}{0.2cm}
\scriptsize
\centering
\caption{PSNR/SSIM results by different noise models.}
\label{tableDNDmodel}
\begin{tabularx}  {\linewidth} { @{} p{3.0cm} C X @{} }
\toprule
Method        & DND~\cite{dataset_Plotz2017BenchmarkingDA}  & Nam~\cite{mention_Nam2016AHA}   \\
\midrule
CBDNet(G)  & 32.52 / 0.79 & 37.62 / 0.9290  \\
CBDNet(HG) & 33.70 / 0.9084 & 38.40 / 0.9453   \\
CBDNet(G+ISP) & 37.41 / 0.9353 & 39.03 / 0.9563 \\
CBDNet(HG+ISP)   & \textbf{37.57} / \textbf{0.9360} & 39.20 / 0.9579 \\
CBDNet(JPEG)   & ---  & \textbf{40.51} / \textbf{0.9745} \\
\toprule
\end{tabularx}
\centering
\vspace{-0.5cm}
\end{table}

\begin{figure*}[!t]
\setlength{\abovecaptionskip}{0.2cm}
\centering
\subfloat{
\begin{minipage}[t]{0.24\textwidth}
\centering
\includegraphics[width=1\textwidth]{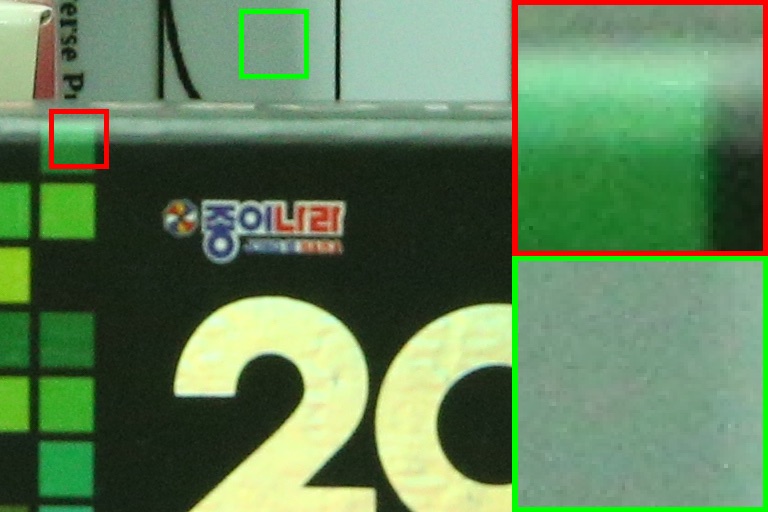}
{\footnotesize  (a) Noisy image}
\end{minipage}\hspace{0.1cm}

\begin{minipage}[t]{0.24\textwidth}
\centering
\includegraphics[width=1\textwidth]{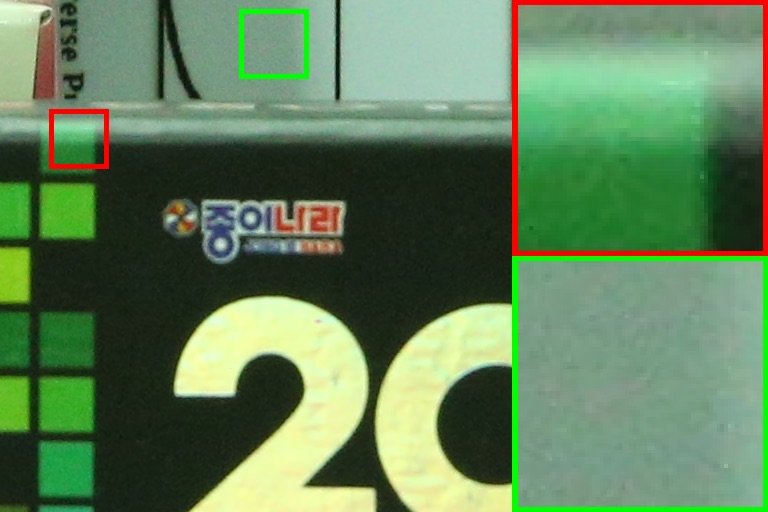}
{\footnotesize  (b) WNNM~\cite{compare_Gu2014WeightedNN}}
\end{minipage}\hspace{0.1cm}

\begin{minipage}[t]{0.24\textwidth}
\centering
\includegraphics[width=1\textwidth]{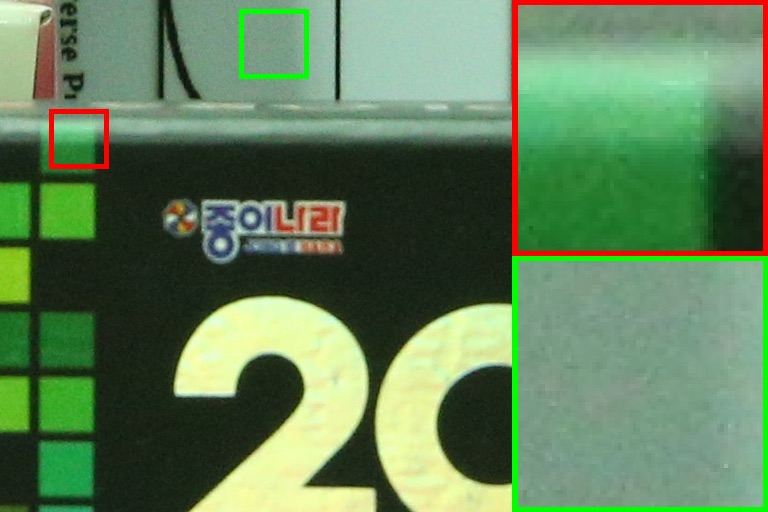}
{\footnotesize  (c) CDnCNN-B~\cite{mention_Zhang2017BeyondAG}}
\end{minipage}\hspace{0.1cm}

\begin{minipage}[t]{0.24\textwidth}
\centering
\includegraphics[width=1\textwidth]{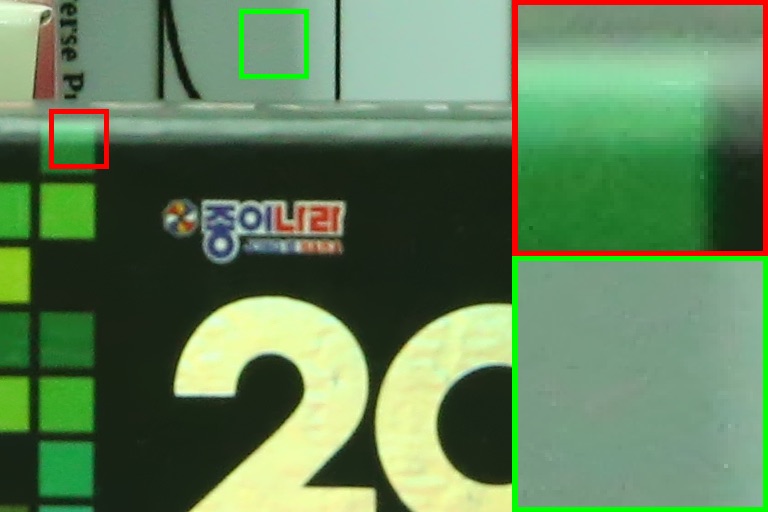}
{\footnotesize  (d) NC~\cite{dataset_Lebrun2015TheNC}}
\end{minipage}
}

\vspace{-3mm}

\subfloat{
\begin{minipage}[t]{0.24\textwidth}
\centering
\includegraphics[width=1\textwidth]{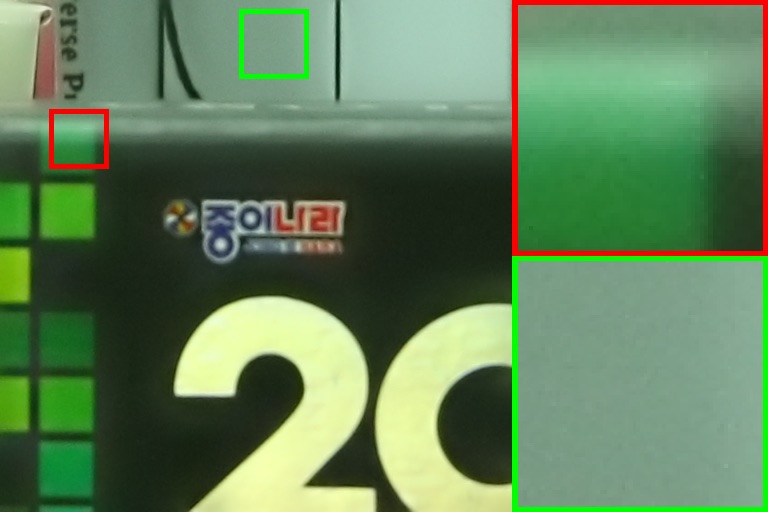}
{\footnotesize  (e) NI~\cite{compare_NI}}
\end{minipage}\hspace{0.1cm}

\begin{minipage}[t]{0.24\textwidth}
\centering
\includegraphics[width=1\textwidth]{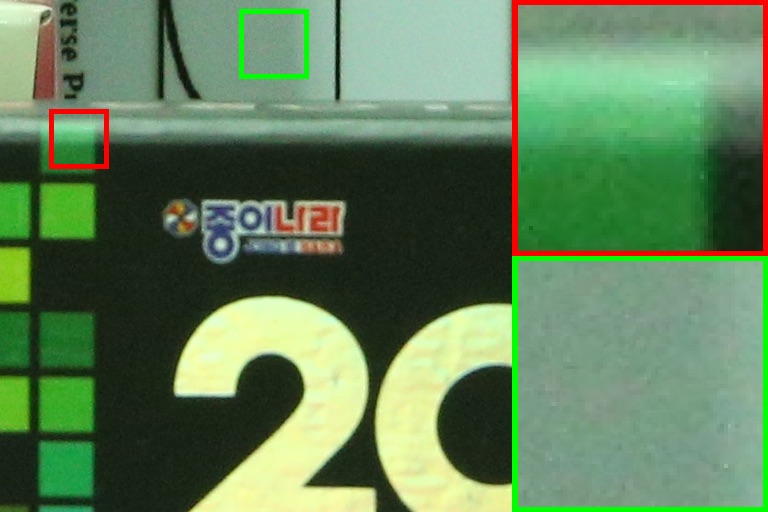}
{\footnotesize  (f) MCWNNM~\cite{Xu2017MultichannelWN}}
\end{minipage}\hspace{0.1cm}

\begin{minipage}[t]{0.24\textwidth}
\centering
\includegraphics[width=1\textwidth]{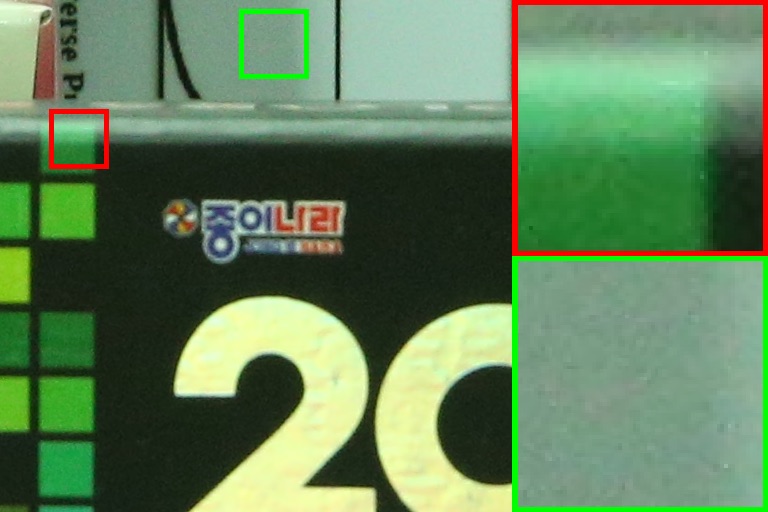}
{\footnotesize  (g) TWSC~\cite{Xu2018ATW}}
\end{minipage}\hspace{0.1cm}

\begin{minipage}[t]{0.24\textwidth}
\centering
\includegraphics[width=1\textwidth]{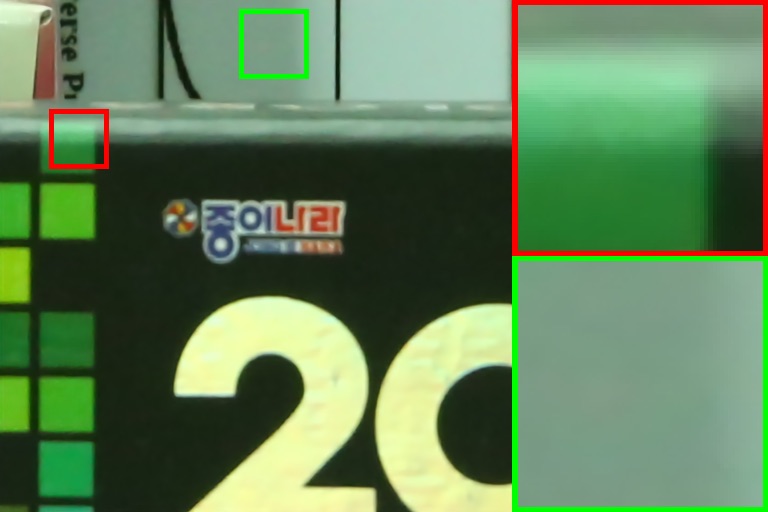}
{\footnotesize  (h) CBDNet}
\end{minipage}
}

\caption{Denoising results of a Nam image by different methods.}
\label{fignamresult}
\end{figure*}

\vspace{-0.4cm}
\paragraph{Nam.} The quantitative and qualitative results are given in Table~\ref{tableNamresults} and Fig.~\ref{fignamresult}.
CBDNet(JPEG) performs much better than CBDNet (\ie, $\sim 1.3$ dB by PSNR) and achieves the best performance in comparison to state-of-the-arts.

%
%
\begin{figure*}[!htbp]
\setlength{\abovecaptionskip}{0.2cm}
\centering
\subfloat{
\begin{minipage}[t]{0.157\textwidth}
\centering
\includegraphics[width=0.97\textwidth]{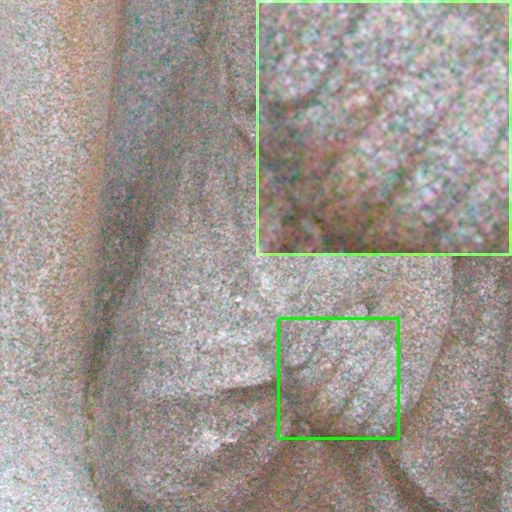}
\end{minipage}\hspace{0.1cm}

\begin{minipage}[t]{0.157\textwidth}
\centering
\includegraphics[width=0.97\textwidth]{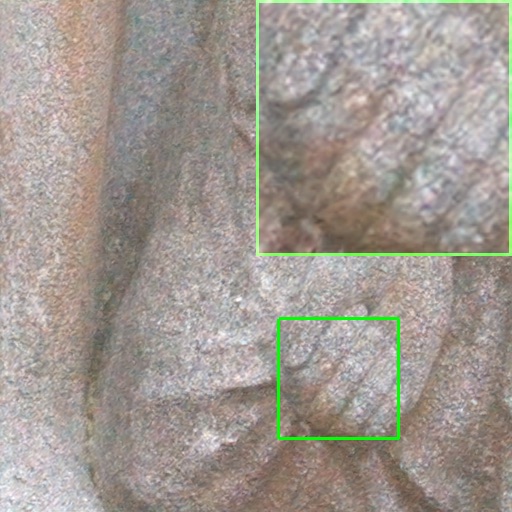}
\end{minipage}\hspace{0.1cm}

\begin{minipage}[t]{0.157\textwidth}
\centering
\includegraphics[width=0.97\textwidth]{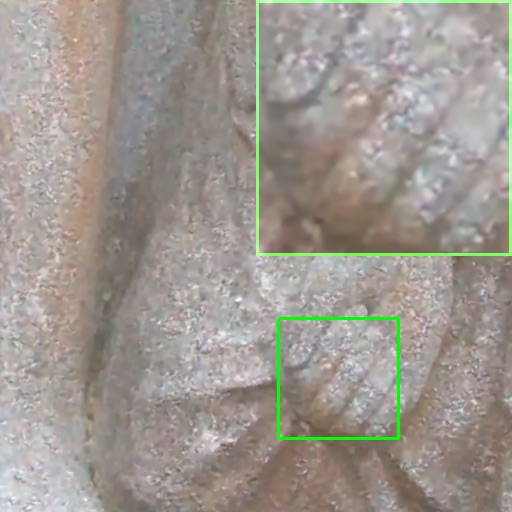}
\end{minipage}\hspace{0.1cm}

\begin{minipage}[t]{0.157\textwidth}
\centering
\includegraphics[width=0.97\textwidth]{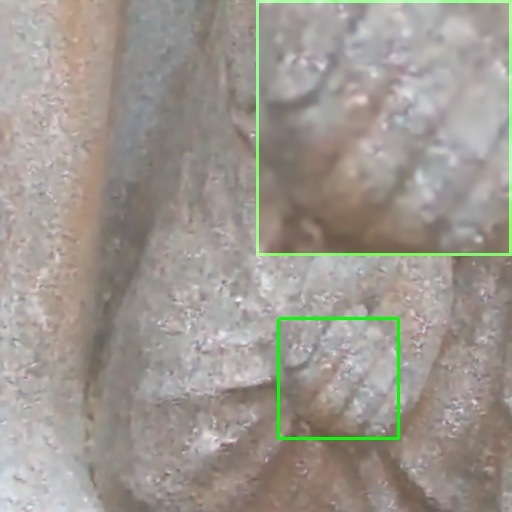}
\end{minipage}\hspace{0.1cm}
\begin{minipage}[t]{0.157\textwidth}
\centering
\includegraphics[width=0.97\textwidth]{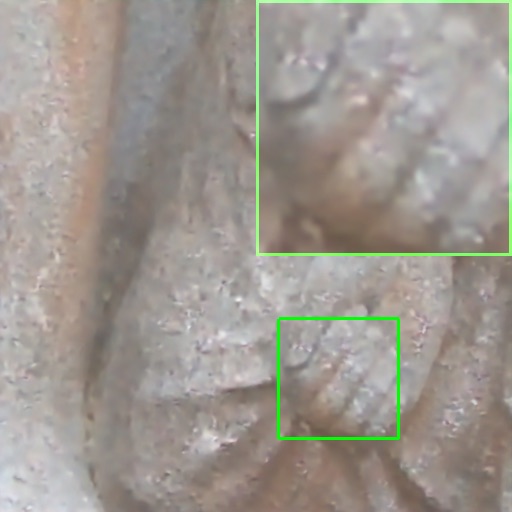}
\end{minipage}\hspace{0.1cm}

\begin{minipage}[t]{0.157\textwidth}
\centering
\includegraphics[width=0.97\textwidth]{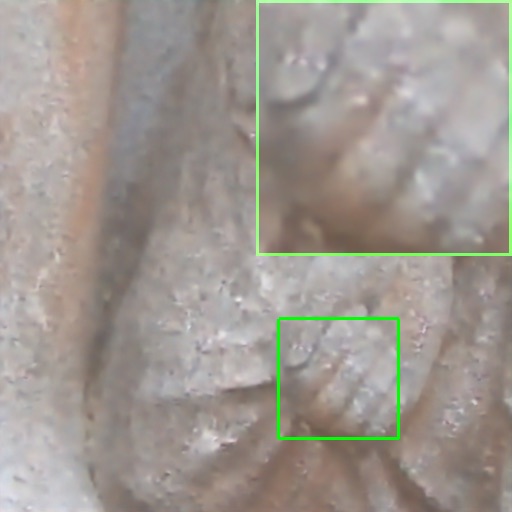}
\end{minipage}
}

\vspace{-3mm}

\subfloat{
\begin{minipage}[t]{0.157\textwidth}
\centering
\includegraphics[width=0.97\textwidth]{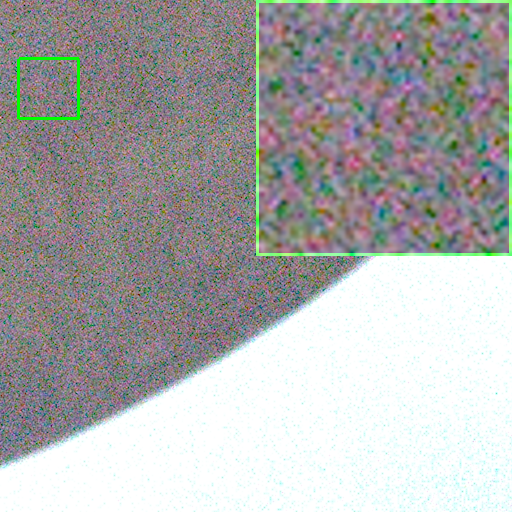}
{\footnotesize  (a) Noisy}
\end{minipage}\hspace{0.1cm}

\begin{minipage}[t]{0.157\textwidth}
\centering
\includegraphics[width=0.97\textwidth]{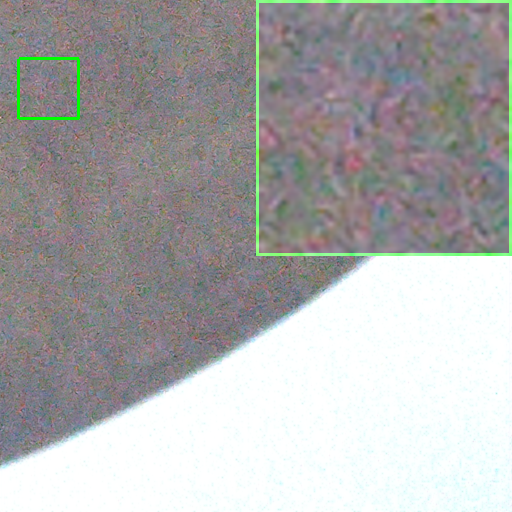}
{\footnotesize  (b) $\gamma = 0.4$}
\end{minipage}\hspace{0.1cm}

\begin{minipage}[t]{0.157\textwidth}
\centering
\includegraphics[width=0.97\textwidth]{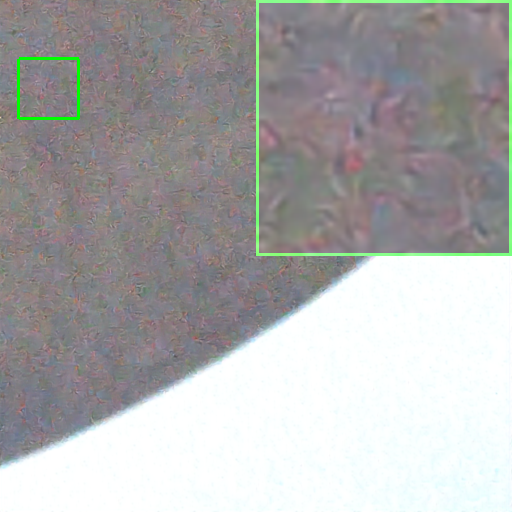}
{\footnotesize  (c) $\gamma = 0.7$}
\end{minipage}\hspace{0.1cm}

\begin{minipage}[t]{0.157\textwidth}
\centering
\includegraphics[width=0.97\textwidth]{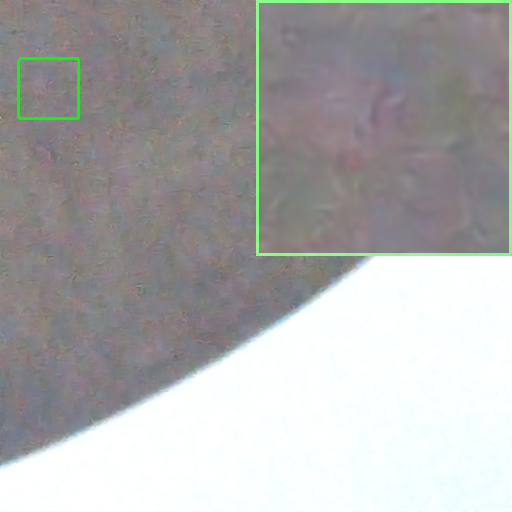}
{\footnotesize  (d) $\gamma = 1.0$}
\end{minipage}\hspace{0.1cm}

\begin{minipage}[t]{0.157\textwidth}
\centering
\includegraphics[width=0.97\textwidth]{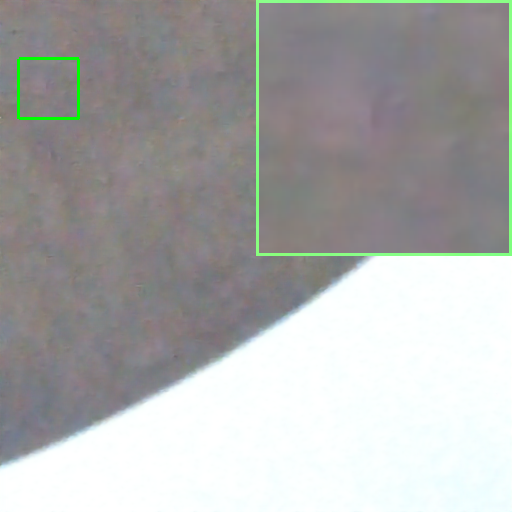}
{\footnotesize  (e) $\gamma = 1.3$}
\end{minipage}\hspace{0.1cm}

\begin{minipage}[t]{0.157\textwidth}
\centering
\includegraphics[width=0.97\textwidth]{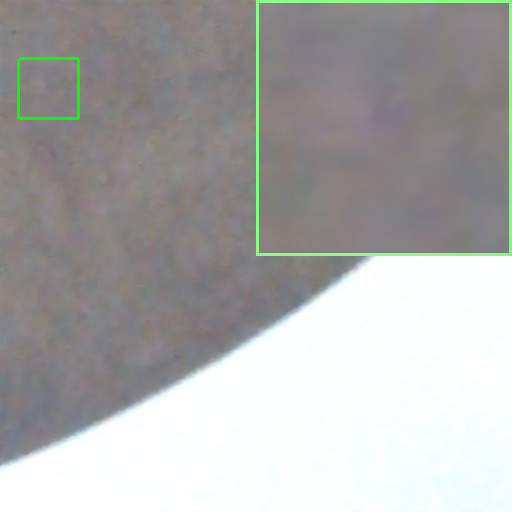}
{\footnotesize  (f) $\gamma = 1.6$}
\end{minipage}
}
\caption{Results by interactive image denoising on two DND images.}
\label{figinteractive}
\vspace{-0.4cm}
\end{figure*}

\subsection{Ablation Studies}

\vspace{-0.1cm}
\paragraph{Effect of noise model.}
Instead of AWGN, we consider heterogeneous Gaussian (HG) and in-camera processing (ISP) pipeline for modeling image noise.
On DND and Nam, we implement four variants of noise models: (i) Gaussian noise (CBDNet(G)), (ii) heterogeneous Gaussian (CBDNet(HG)), (iii) Gaussian noise and ISP (CBDNet(G+ISP)), and (iv) heterogeneous Gaussian and ISP (CBDNet(HG+ISP), \ie, full CBDNet.
For Nam, CBDNet(JPEG) is also included.
Table \ref{tableDNDmodel} shows the PSNR/SSIM results of different noise models.

\vspace{-0.5cm}
\paragraph{G vs HG.} Without ISP, CBDNet(HG) achieves about $0.8\sim1$ dB gain over CBDNet(G).
When ISP is included, the gain by HG is moderate, \ie, CBDNet(HG+ISP) only outperforms CBDNet(G+ISP) about $0.15$ dB.

\vspace{-0.5cm}
\paragraph{w/o ISP.} In comparison, ISP is observed to be more critical for modeling real image noise.
In particular, CBDNet(G+ISP) outperforms CBDNet(G) by $4.88$ dB, while CBDNet(HG+ISP) outperforms CBDNet(HG) by $3.87$ dB on DND.
For Nam, the inclusion of JPEG compression in ISP further brings a gain of 1.31 dB.

\begin{figure}[!h]
\setlength{\abovecaptionskip}{0.2cm}
\centering
\begin{minipage}[t]{0.22\textwidth}
\centering
\includegraphics[width=0.96\textwidth]{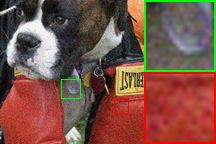}
{\footnotesize  (a) Noisy image}
\end{minipage}\hspace{0.25cm}
\begin{minipage}[t]{0.22\textwidth}
\centering
\includegraphics[width=0.96\textwidth]{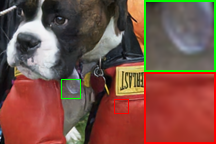}
{\footnotesize  (b) CBDNet(Syn)}
\end{minipage}

\begin{minipage}[t]{0.22\textwidth}
\centering
\includegraphics[width=0.96\textwidth]{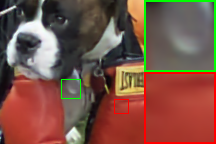}
{\footnotesize  (c) CBDNet(Real)}
\end{minipage}\hspace{0.25cm}
\begin{minipage}[t]{0.22\textwidth}
\centering
\includegraphics[width=0.96\textwidth]{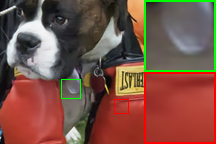}
{\footnotesize  (d) CBDNet(All)}
\end{minipage}

\caption{\small Denoising results of CBDNet trained by different data.}
\label{figtrainingstrategy}
\vspace{-0.5cm}
\end{figure}

\vspace{-0.5cm}
\paragraph{Incorporation of synthetic and real images.}\label{Mixlearningstrategy}
We implement two baselines: (i) CBDNet(Syn) trained only on synthetic images, and (ii) CBDNet(Real) trained only on real images, and  rename our full CBDNet as CBDNet(All).
Fig. \ref{figtrainingstrategy} shows the denoising results of these three methods on a NC12 image.
Even trained on large scale synthetic image dataset, CBDNet(Syn) still cannot remove all real noise, partially due to that real noise cannot be fully characterized by the noise model.
CBDNet(Real) may produce over-smoothing results, partially due to the effect of imperfect noise-free images.
In comparison, CBDNet(All) is effective in removing real noise while preserving sharp edges.
Also quantitative results of the three models on DND are shown in Table \ref{tableDNDresults}.
CBDNet(All) obtains better PSNR/SSIM results than CBDNet(Syn) and CBDNet(Real).

\begin{figure}
\setlength{\abovecaptionskip}{0.2cm}
\centering
\subfloat{
    \begin{minipage}[b]{0.215\textwidth}
    \centering
        		\includegraphics[width=1\textwidth]{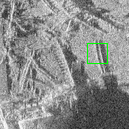}
		\footnotesize{(a) Noisy image}
    \end{minipage}
    \hspace{0.25cm}
   \begin{minipage}[b]{0.215\textwidth}
    \centering
		\begin{overpic}[width=1\textwidth]{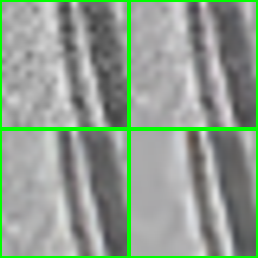}
		\put(1,52){\color{white}{\footnotesize noisy}}
		\put(52,52){\color{white}{\footnotesize $\alpha=0.5$}}
		\put(1,2){\color{white}{\footnotesize $\alpha=0.4$}}
		\put(52,2){\color{white}{\footnotesize $\alpha=0.3$}}
		\end{overpic}
		\footnotesize{(b) Denoised patches}
    \end{minipage}
    }

\caption{\small Denoising results of CBDNet with different $\alpha$ values}
\label{figasyloss} 
\vspace{-0.5cm}
\end{figure}

\vspace{-0.5cm}
\paragraph{Asymmetric loss.}
Fig.~\ref{figasyloss} compares the denoising results of CBDNet with different $\alpha$ values, \ie, $\alpha = 0.5, 0.4$ and $0.3$.
CBDNet imposes equal penalty to under-estimation and over-estimation errors when $\alpha = 0.5$, and more penalty is imposed on under-estimation error when $\alpha < 0.5$.
It can be seen that smaller $\alpha$ (\ie, $0.3$) is helpful in improving the generalization ability of CBDNet to unknown real noise.

\subsection{Interactive Image Denoising}\label{secinteractive}
Given the estimated noise level map $\hat{\sigma}(\mathbf{y})$, we introduce a coefficient $\gamma \, (>0)$ to interactively modify $\hat{\sigma}(\mathbf{y})$ to $\hat{\varrho} = \gamma \cdot \hat{\sigma}(\mathbf{y})$.
By allowing the user to adjust $\gamma$, the non-blind denoising subnetwork takes $\hat{\varrho}$ and the noisy image as input to obtain denoising result.
Fig.~\ref{figinteractive} presents two real noisy DND images as well as the results obtained using different $\gamma$ values.
By specifying $\gamma = 0.7$ to the first image and $\gamma = 1.3$ to the second, CBDNet can achieve the results with better visual quality in preserving detailed textures and removing sophisticated noise, respectively.
Such interactive scheme can thus provide a convenient means for adjusting the denosing results in practical scenario.

\section{Conclusion}
We presented a CBDNet for blind denoising of real-world noisy photographs.
The main findings of this work are two-fold.
First, realistic noise model, including heterogenous Gaussian and ISP pipeline, is critical in making the learned model from synthetic images be applicable to real-world noisy photographs.
Second, the denoising performance of a network can be boosted by incorporating both synthetic and real noisy images in training.
Moreover, by introducing a noise estimation subnetwork into CBDNet, we were able to utilize asymmetric loss to improve its generalization ability to real-world noise, and perform interactive denoising conveniently.

\section{Acknowledgements}
This work is supported by NSFC (grant no. 61671182, 61872118, 61672446) and HK RGC General Research Fund (PolyU 152216/18E).

\clearpage
{\small
\bibliographystyle{ieee}
\bibliography{egbib}
}

\end{document}